\documentclass[letterpaper, 10 pt, conference]{ieeeconf}
\IEEEoverridecommandlockouts

\usepackage{tikz}
\usepackage{textcomp}
\usepackage{hyperref}
\usepackage{lipsum}

\newcommand\copyrighttext{%
  \footnotesize \textcopyright 2022 IEEE. Personal use of this material is permitted.
  Permission from IEEE must be obtained for all other uses, in any current or future 
  media, including reprinting/republishing this material for advertising or promotional 
  purposes, creating new collective works, for resale or redistribution to servers or 
  lists, or reuse of any copyrighted component of this work in other works.}
\newcommand\copyrightnotice{%
\begin{tikzpicture}[remember picture,overlay]
\node[anchor=south,yshift=10pt] at (current page.south) {\fbox{\parbox{\dimexpr\textwidth-\fboxsep-\fboxrule\relax}{\copyrighttext}}};
\end{tikzpicture}%
}

\title{\LARGE \bf Analysis of Randomization Effects on Sim2Real Transfer in Reinforcement Learning for Robotic Manipulation Tasks}

\author{Josip Josifovski$^{1*}$, Mohammadhossein Malmir$^{1}$, Noah Klarmann$^{2}$, Bare Luka \v{Z}agar$^{1}$,\\ Nicol\'as Navarro-Guerrero$^{3}$ and Alois Knoll$^{1}$
\thanks{$*$ Corresponding Author: Josip Josifovski {\tt\footnotesize josip.josifovski@tum.de}}%
\thanks{$^{1}$ Department of Informatics, Technical University of Munich, Germany.}%
\thanks{$^{2}$ Rosenheim University of Applied Sciences, Rosenheim, Germany.}%
\thanks{$^{3}$ Deutsches Forschungszentrum f\"ur K\"unstliche Intelligenz GmbH, Germany.}%
\thanks{This work has been financially supported by AI4DI project, which has received funding from the ECSEL Joint Undertaking (JU) under grant agreement No.\ 826060. The JU receives support from the European Union’s Horizon 2020 research and innovation programme, Germany, Austria, Czech Republic, Italy, Latvia, Belgium, Lithuania, France, Greece, Finland, and Norway.}}

\usepackage{cite}  
\usepackage{amsmath,amssymb,amsfonts}
\usepackage{algorithmic}
\usepackage{graphicx}
\usepackage{textcomp}
\usepackage{xcolor}
\usepackage{xfrac}

\definecolor{noteyellow}{RGB}{255,186,0}	 
\definecolor{notegreen}{RGB}{0,128,0}        
\definecolor{noteblue}{RGB}{0,0,205}         
\definecolor{noteviolet}{RGB}{199,21,133}    

\usepackage{url}
\usepackage[protrusion=true,expansion=false]{microtype}
\usepackage{hyperref}
\hypersetup{
  pdftitle={Analysis of Randomization Effects on Sim2Real Transfer in Reinforcement Learning for Robotic Manipulation Tasks },
  pdfauthor={Josifovski, Malmir, Klarmann, \v{Z}agar, Navarro-Guerrero, and Knoll},
  pdfsubject={Robotics (cs.RO); Artificial Intelligence (cs.AI)}
  pdfkeywords={sim2real, Reality Gap, Reinforcement Learning, Sequential Randomization, Real-World Robotic Tasks},%
  colorlinks=true,
  linktoc=all,
  linkbordercolor=white,
  linkcolor=black,
  citecolor=black,
  urlcolor=black,
}

\newcommand{\specialcell}[2][c]{%
  \begin{tabular}[#1]{@{}c@{}}#2\end{tabular}}

\begin{document}

\maketitle
\copyrightnotice
\thispagestyle{empty}
\pagestyle{empty}

\begin{abstract}
Randomization is currently a widely used approach in Sim2Real transfer for data-driven learning algorithms in robotics. Still, most Sim2Real studies report results for a specific randomization technique and often on a highly customized robotic system, making it difficult to evaluate different randomization approaches systematically. To address this problem, we define an easy-to-reproduce experimental setup for a robotic reach-and-balance manipulator task, which can serve as a benchmark for comparison. We compare four randomization strategies with three randomized parameters both in simulation and on a real robot. Our results show that more randomization helps in Sim2Real transfer, yet it can also harm the ability of the algorithm to find a good policy in simulation. Fully randomized simulations and fine-tuning show differentiated results and translate better to the real robot than the other approaches tested.  
\end{abstract}


\section{Introduction}
Current research in reinforcement learning (RL) shows impressive results in disembodied environments, where RL agents outperform humans in video games \cite{Mnih2013Playinga}, Go \cite{Silver2016Mastering}, or develop creative strategies for multi-agent hide-and-seek scenarios \cite{Baker2020Emergent}. 
However, applying such models to control real robots is more challenging. For example, real-world training of a data-driven model \cite{Levine2018Learning} can be impractical or unsafe in many cases where environments are dynamic or computational, and hardware resources are limited, which is often the case in robotics. 
Simulations seem like a viable alternative for training robots -- with the ever-increasing simulation accuracy and computational speedup, virtually unlimited training data is available to train RL agents for robotic tasks. 
There are many simulation platforms \cite{Brockman2016OpenAI}, \cite{Yu2020MetaWorld} \cite{Makoviychuk2021Isaac}, \cite{Falotico2017Connecting}, \cite{Todorov2012MuJoCo} that target robotic simulations both in research and in industrial context. 
However, simulations are always an approximation of the real environment. Thus, models trained in simulation often fail in the real world, or their performance decreases considerably, which is known as the reality gap problem.  

Currently, two predominant approaches for minimizing the reality gap are developing highly realistic simulations or using imperfect simulations in combination with randomization strategies \cite{Malmir2020Robust}. 
Developing highly realistic simulations depends on processes like system identification, which are often hardware-specific, and the parameters identified with such processes might vary significantly based on conditions in the environment \cite{Zhao2020SimtoReal}. 
Furthermore, it is not always easy to determine which real-world phenomena and to what level of precision should be modeled in highly realistic simulations.

On the contrary, randomization strategies take advantage of the generalization capabilities of neural networks and randomize aspects of the task that are challenging to measure or model precisely \cite{Tobin2017Domain}. The general assumption, in this case, is that the actual real-world parameters are covered in the randomization ranges or that models trained on randomized data learn more robust policies. 
Arguably, randomization strategies are more applicable in general than learning on highly realistic simulations because the policies learned under randomization might adapt and work in dynamic environments or settings not encountered during training. 

However, randomization strategies come at the cost of a more significant number of training samples to learn a task. 
Moreover, randomization can lead to sub-optimal solutions or even prevent learning, which can be attributed to the inherent unpredictability of the randomized simulations. 
Therefore, it is essential to quantify the effects of randomization on the learning of policies that would transfer best to the real world. Such systematic studies are still lacking in the field -- while there are many successful examples of Sim2Real transfer, they are usually a one-time implementation on a highly-customized system, which hinders experiment reproducibility, benchmarking, or comparative analysis of Sim2Real transfer approaches.

Thus, here we present a comparative analysis of training strategies with different levels of randomization for a robotic reaching task. 
The chosen task is sufficiently simple to train RL models quickly, while it is still complex enough to capture the effects of the reality gap. Moreover, a deterministic solution for the task exists and can be used as a baseline. 
The simulation environments and code, together with a demonstration video, have been made available to encourage reproducibility at https://josifovski.github.io/sim2real-randomization-effects/. 
\section{Background and Related Work}

Training in simulation for robotic control and the reality gap problem have been addressed in many earlier research studies \cite{Jakobi1995Noise}, \cite{Jakobi1998Running}, \cite{Zagal2004Back},\cite{Koos2013Transferability}. More recently, with the rise of deep learning and the advances in computation, Sim2Real transfer became a common practice in fields like computer vision \cite{Tobin2017Domain}, \cite{Josifovski2018Object}, where synthetic data can be used to train models in simulation completely or to improve their performance when combined with real data. In the last few years, bringing reinforcement learning in combination with deep learning and improved simulations opened the possibility of using Sim2Real transfer effectively in the robotics domain. Moreover, it allowed simulated experiences to replace some of the hard-to-acquire robotic experiences on a real robot. A recent overview of the Sim2Real approaches in robotics is presented by Zhao et al.\ \cite{Zhao2020SimtoReal}. 

Some of the Sim2Real approaches aim to precisely reproduce the relevant real-world phenomena in simulation \cite{James20163D}, \cite{Kaspar2020Sim2Real}. For instance, James and Johns \cite{James20163D} trained a 7-DOF robotic arm in simulation for locating and grasping a cube by adjusting the simulation to resemble the real world as closely as possible. Similarly, Kaspar et al.\  \cite{Kaspar2020Sim2Real} show successful Sim2Real transfer for a peg in-hole task with a 7-DOF robotic arm by training only in simulation, where they perform system identification to resemble the real robot's dynamics in simulation accurately. 

As matching and correctly simulating all relevant real-world phenomena for a robotic task is often complicated to achieve, other approaches \cite{Peng2018SimtoReal}, \cite{Tobin2017Domain}, \cite{OpenAI2019Solving}, \cite{James2019SimtoReal}
rely on randomization of the critical simulation parameters during model training to increase model robustness and minimize the effects of the mismatch between the simulation and the real world. For example, Peng et al.\ \cite{Peng2018SimtoReal} trained a robotic arm to perform a pushing task by randomizing the robot's physical properties and the simulation's dynamics.
Similarly, Tobin et al.\ \cite{Tobin2017Domain} employed domain randomization to create millions of unrealistically-appearing training images for robust object localization in a robotic grasping task. 
In the same direction, James et al. \cite{James2019SimtoReal} combined a domain randomization technique with a generative model to facilitate the learning to grasp unknown objects with a robotic arm. 
OpenAI \cite{OpenAI2019Solving} successfully transferred policies learned only in simulation to a real robotic hand to solve a Rubic's cube. They achieved this by automated domain randomization, a form of curriculum learning where the randomization ranges become increasingly wider to make the model learn more general solutions instead of overfitting to a specific randomization seen during training. An overview of randomization techniques in robot learning for Sim2Real transfer was presented by Muratore et al.\ \cite{Muratore2022Robot}.

Many Sim2Real approaches employ domain adaptation, where the model trained in simulation is fine-tuned on the real robot as a final step to improve its performance. The domain adaptation can also be framed as a continual learning problem by considering the same robotic task in simulation and the real world as two separate tasks to solve. 
For example, Rusu et al.\ \cite{Rusu2017SimtoReal} trained a model for a reaching task in simulation and then reused the model's features to train a second model in the real world. They reported that due to the reusability of the skills learned in simulation, the additional training steps in the real world are drastically decreased.

While domain adaptation is a viable last step, minimizing the time needed to train on a real robot or circumvent it entirely is always important. Thus, it is necessary to measure how different algorithms' properties or training strategies influence the difficulty of learning a task in simulation and how the performance in simulation relates to the performance on a real robot. In this direction,  Kadian et al.\ \cite{kadian2020sim2real} developed a complex setup of a real environment and 3D scan it to create a virtual replica to measure Sim2Real predictivity for a visual navigation robotic task.

Similarly, here we use a real environment and a virtual replica to quantify the Sim2Real transfer. We also describe how well the simulation performance predicts the performance in a real environment. Unlike Kadian et al. \cite{kadian2020sim2real}, we focus on randomization strategies and their effects on learning a robotic manipulator task, with an accent on environment simplicity in order to mitigate confounding effects and ease experiment reproducibility.

\section{Methodology}

\begin{figure*}[htbp]
	 \centering
	 \includegraphics[scale = 0.89]{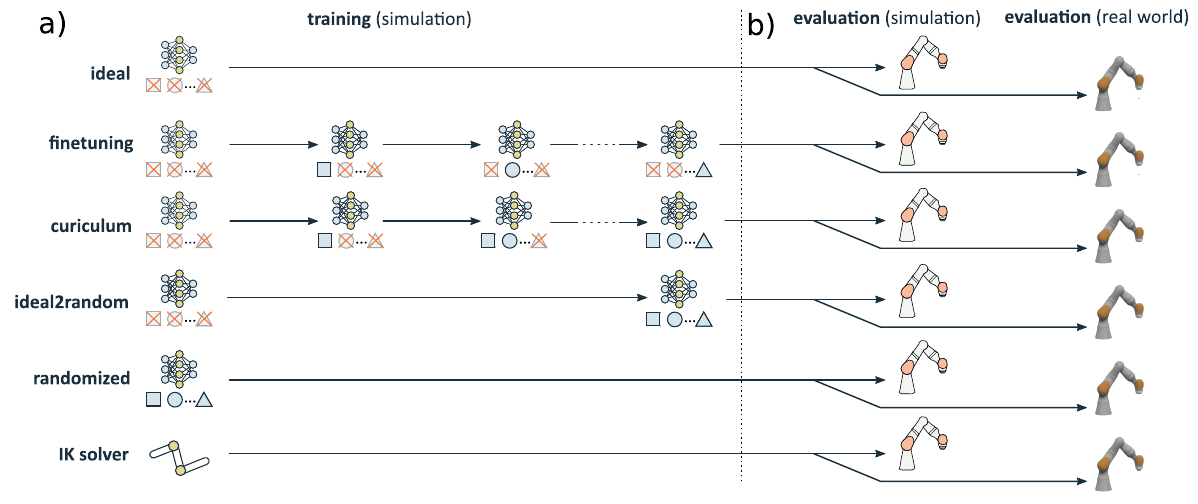}
	 \caption{Methodology a) Different training strategies. Each shape represents a different randomization parameter and whether it is enabled or not during training. b) The simulation-trained model is evaluated under equivalent conditions in simulation and on the real robot.}
\label{fig:Approach}
\end{figure*}

We first formally specify a robotic manipulation task suitable for evaluating Sim2Real transfer. 
Then we describe the randomization strategies used. 
Finally, we explain the evaluation procedure in both simulation and real-world setups. 

\subsection{Task Formulation}
The task is to learn the differential inverse kinematics of a 2-joint robot, always starting from the same `home' position -- the robot is upright with both joints at 0 degrees rotation. The end-effector should be moved to a specific reachable target position by continuously controlling the joint velocities. 
Typical RL reaching benchmarks and implementations (e.g., \cite{OpenAIgymReacher, Navarro-Guerrero2017Improving, Stahlhut2015Interactiona}) only consider the goal of reaching the target location. However, there is no consideration for what happens after achieving the goal, which poses potential safety issues when transferring the behavior to the real world. Thus, we require the agent to learn to reach and stay at the specified target until the end of the episode. We call this the \textit{reach-and-balance} task:
\begin{equation}
\mathbf{s}_t = [\Delta{x}_t, \Delta{y}_t, \Delta{z}_t,{q}_1,{q}_2] \in \mathbb{R}^5
\end{equation}

\noindent $s_t$ is the state of the agent at a specific timestep, $\Delta{x}$, $\Delta{y}$ and $\Delta{z}$ are the distances between the end-effector and the target along the corresponding coordinate axis, ${q_1}$ and ${q_2}$ are the joint positions. 
The actions the agent can perform are specified by:
\begin{equation}
\mathbf{a}_t = [\dot {q}_1^{\prime},\dot{q}_2^{\prime}] \in \mathbb{R}^2
\end{equation}

\noindent where $\dot {q_i}^{\prime}$ are the angular velocities for the first and the second joint starting from the robot's home position.
The reward function is defined as:
\begin{equation}
\mathbf{r}_t = 
    \begin{cases}
    - \mathbf{d}_t   \ & if\ C\ is\ false\ \\
    - \mathbf{d}_t * (\mathbf{T} - \mathbf{t}) & otherwise
    \end{cases}
    \in \mathbb{R}
\end{equation}

\noindent where $\mathbf{d}_t$ is the Euclidean distance between the target and the end-effector, $\mathbf{T}$ is the episode duration, and $\mathbf{\textit{C}}$ is the terminal condition. During training, $\mathbf{\textit{C}}$ becomes true when the robot collides with the floor. However, during evaluation, $\mathbf{\textit{C}}$ becomes true if any of the joints reach a predefined safety rotation limit. The goal of the agent is to maximize the episode reward: 
\begin{equation}
\mathbf{R} = \sum_{t=1}^{T} \mathbf{r}_t 
\end{equation}

\subsection{Training Strategies}
To investigate randomization effects, we selected different randomization strategies based on their conceptual differences, see Figure \ref{fig:Approach}:

\begin{itemize}
    \item \textbf{Ideal simulation} -- The simulation environment does not include any randomization parameter relevant for Sim2Real transfer.

	\item \textbf{Fine-tuning} -- It uses a model pre-trained on an ideal simulation as the starting point. The model is further trained sequentially for each randomization parameters of interest, one at a time. 

	\item \textbf{Curriculum learning} -- It uses a model pre-trained on an ideal simulation as the starting point. The model is further trained sequentially with an increasing number of randomized parameters.

	\item \textbf{ideal2randomized} -- It uses a model pre-trained on an ideal simulation as the starting point. The model is further trained with all the randomized parameters at once. 
	
	\item \textbf{Randomized simulation} -- A model is trained from scratch, with all the randomization parameters at once. 

	\item \textbf{Inverse kinematics} -- We use an inverse kinematics solver as a baseline. It is a deterministic strategy. 
\end{itemize}

\section{Experiment Setup}
The simulation environment \cite{Josifovski2020Continual} is developed with Unity \cite{Unity} and has a comparable API to OpenAI Gym \cite{Brockman2016OpenAI}. It allows parallel simulation of many robots at once to speed-up the training process. 
We used the KUKA LBR iiwa 14 robot manipulator \cite{Kuka} due to its ubiquity in both the research and industrial domain. The simulated robot parameters and meshes are obtained from the URDF data provided by ROS-Industrial \cite{ROSindustrial}. 
The real robot is controlled via ROS using the IIWA stack \cite{Hennersperger2017MRIBased}. 

\subsection{Randomization Parameters}
\label{randomization_params}
We select three parameters that are commonly used in Sim2Real transfer in robotics \cite{Peng2018SimtoReal} \cite{Tan2018SimtoReal}, \cite{Zhao2020SimtoReal}, i.e., Latency ($L$), Torque ($T$), and Noise ($N$). 

\textbf{Latency ($L$):} In an ideal simulation condition, there is no latency. For the randomized case, inspired by Tan et al.\ \cite{Tan2018SimtoReal}, we collect a list of ideal observations for every timestep. Then, a random latency value is drawn from a uniform distribution. 
A linear interpolation of the two consecutive ideal observations closest to the timestep ($t'$) resulting from the current timestep ($t$) minus the latency is used to determine the latency-affected state of the agent.
If the resulting state ($s_{t'}$) precedes the latency-affected state from the previous timestep ($s_{t'-1}$), then the same latency-affected state from $s_{t'-1}$ is used.

\textbf{Torque ($T$):} The joint velocity control is implemented using the spring-damper equation as explained in Unity Robotics Hub \cite{unityRoboticsHub2022}. Here, we randomize the stiffness and damping coefficients for the joints, which results in a randomized torque profile for reaching the commanded velocity, similarly to  Peng et al. \cite{Peng2018SimtoReal}. On the contrary, the commanded joint velocity is reached instantaneously in an ideal simulation condition.

\textbf{Noise ($N$):} Random noise drawn from a uniform distribution is added to simulate sensor reading noise \cite{Tan2018SimtoReal, Peng2018SimtoReal}.

\begin{figure}[htbp]
    \centering
	\includegraphics[width=\columnwidth]{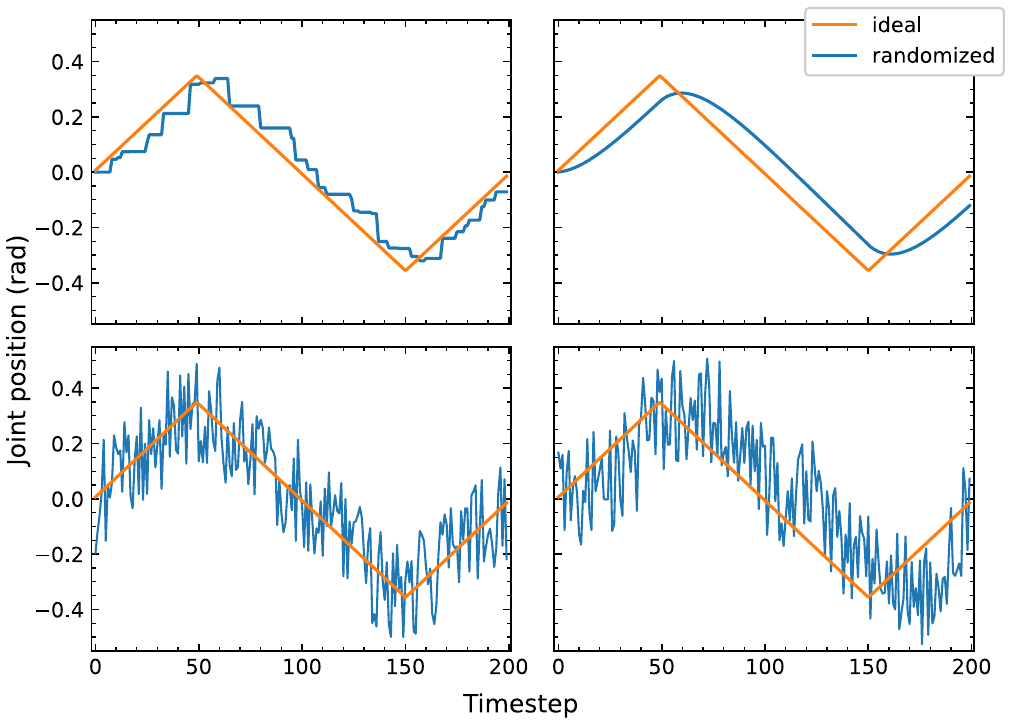}
	\caption{Effects of different randomization parameters on the observation. Plots show the agent's observation for the second joint angle after requesting a fixed 20 deg/sec rotational speed for 1 second clockwise, 2 seconds counter-clockwise, and 1 second clockwise. Top left: randomized latency. Top right: randomized torque. Bottom left: randomized noise. Bottom right: combined randomization of all parameters.}
    \label{fig:randomized_parameters}
\end{figure}

The ranges for the randomization parameters can be defined based on system identification, or unrealistic ranges can be used, expecting that the true system values are included within the range.  
The latter is used because we believe this technique can generalize better and thus more effectively cope with the Sim2Real transfer. 
An agent trained in an ideal simulation was used to determine the range of the randomization parameters. The range of the corresponding parameter was determined by increasing the parameter range until the agent's performance degraded by more than 10\%.
Table \ref{table:randomization_parameter_details} summarizes the values for all three randomization parameters. Figure \ref{fig:randomized_parameters} shows a visual representation of the randomization parameters.

\begin{table}[h]
    \begin{center}
    \caption{Randomization parameters.}
    \label{table:randomization_parameter_details}
    \begin{tabular}{llccc}
    \hline\noalign{\smallskip}
        &    &   \specialcell{Target\\Domain} & \specialcell{Correlation to\\Ideal Obs.} & \specialcell{Effect\\Freq.} \vspace{-10pt}\\
    Param.\ & Range & & &\\
    \hline\noalign{\smallskip}
    $L$ & \specialcell{[0, 1] sec} & obser.  &  linear & timestep  \\
    $T$ & \specialcell{stiffness: [1, 100]\\damping: [1, 100]} & action  &  non-linear & episode  \\
    $N$ & \specialcell{[0, 10] \%} & obser.  &  uncorrelated & timestep  \\
    \hline\noalign{\smallskip}
    \end{tabular}
    \end{center}
\end{table}

\subsection{Training Procedure}
We used Proximal Policy Optimization (PPO) \cite{Schulman2017Proximal} for all the experiments. PPO is commonly used across many different domains, thus making it suitable for comparative analysis. 
Another benefit of PPO is that it uses a conservative clipped objective function that prevents large policy changes between two consecutive updates. Clipped updates are especially useful since the inherent uncertainty in the observations due to randomization might lead to large policy updates based on incorrect observations. 
We use the implementation from StableBaselines \cite{Hill2018Stable}, which allows parallelization of the training using many environments simultaneously. The network architecture for both the Actor and the Critic is a 2-layer MLP with 64 units per layer and a tanh activation function. The hyperparameters for a similar robotic task, the ReacherBulletEnv-v0 by Raffin et al.\ \cite{Raffin2020zoo} were used. However, the clipping range was changed after preliminary test to 0.1 to reduce the variance during randomization training. 
The number of parallel timesteps per update was also proportionally updated from 2048 to 256, as we train on 64 environments in parallel. Table \ref{table:hyperparameters_details} provides a summary of all the relevant training hyperparameters.

\begin{table}[h]
    \begin{center}
    \caption{Training hyperparameters.}
    \label{table:hyperparameters_details}
    \begin{tabular}{lc}
    \hline\noalign{\smallskip}
     Bias-variance trade-off ($\lambda$)  &  0.95 \\
     Discount factor ($\gamma$)   &   0.99 \\
     Learning rate ($\alpha$)  & 0.00025 \\
     Value function coefficient ($c_{1}$)  &  0.5 \\
     Entropy coefficient ($c_{2}$) & 0.0 \\
     Clip range ($\epsilon$)  &  0.1 \\ 
     Number of optimization epochs  &  10 \\ 
     Mini-batches  &  32 \\ 
    \hline\noalign{\smallskip}
    \end{tabular}
    \end{center}
\end{table}

Episode length is 250 timesteps, where one timestep is 20ms of simulation time, which amounts to 5 seconds of simulation time per episode during training with a control frequency of 50 Hz. 

The max joint speed was set to $1 \; \sfrac{rad}{sec}$. However, this value can be exceeded when using the Torque randomization parameter.

All models are trained for a total of 40 million timesteps for five different initializations. For each strategy and initialization, the targets were randomly sampled along the half-sphere around the robot, starting from a height of 10 cm above the floor. While the number of total training steps for each run is the same, the number of training targets might differ since some episodes might terminate earlier due to collisions.

Under the ideal strategy, the model is trained entirely on non-randomized simulation, while for the randomized strategy all parameters are randomized for the whole duration of the training.
For the fine-tuning, curriculum, and ideal2randomized strategies, a model pre-trained for 31 million timesteps under the ideal strategy is used as the starting point and trained for additional 9 million timesteps. This division is based on the approach of determining the randomization ranges described in \ref{randomization_params}, i.e., for the sequential training strategies we use about 10\% of the training time per randomization parameter w.r.t. the training time of the pre-trained ideal model we start from. Under the fine-tuning strategy, the pre-trained model is sequentially trained on three randomized simulations for 3 million timesteps per simulation, where only one randomization parameter is enabled at a time. Similarly, for the curriculum strategy, the pre-trained model is further trained on a sequence of three randomized simulations for 3 million timesteps per simulation, where an additional randomization parameter is enabled each time, leading from less to more randomized training. Under the fine-tuning and curriculum strategies, the parameter randomization sequence might affect the final performance; hence we test all the permutations of the randomization parameters. Lastly, under the ideal2randomized strategy, the pre-trained model on ideal simulation is further trained directly on fully-randomized simulation for 9 million timesteps.

\subsection{Evaluation Procedure}
\begin{figure}[htbp]
    \centering
	\includegraphics[scale = 0.31]{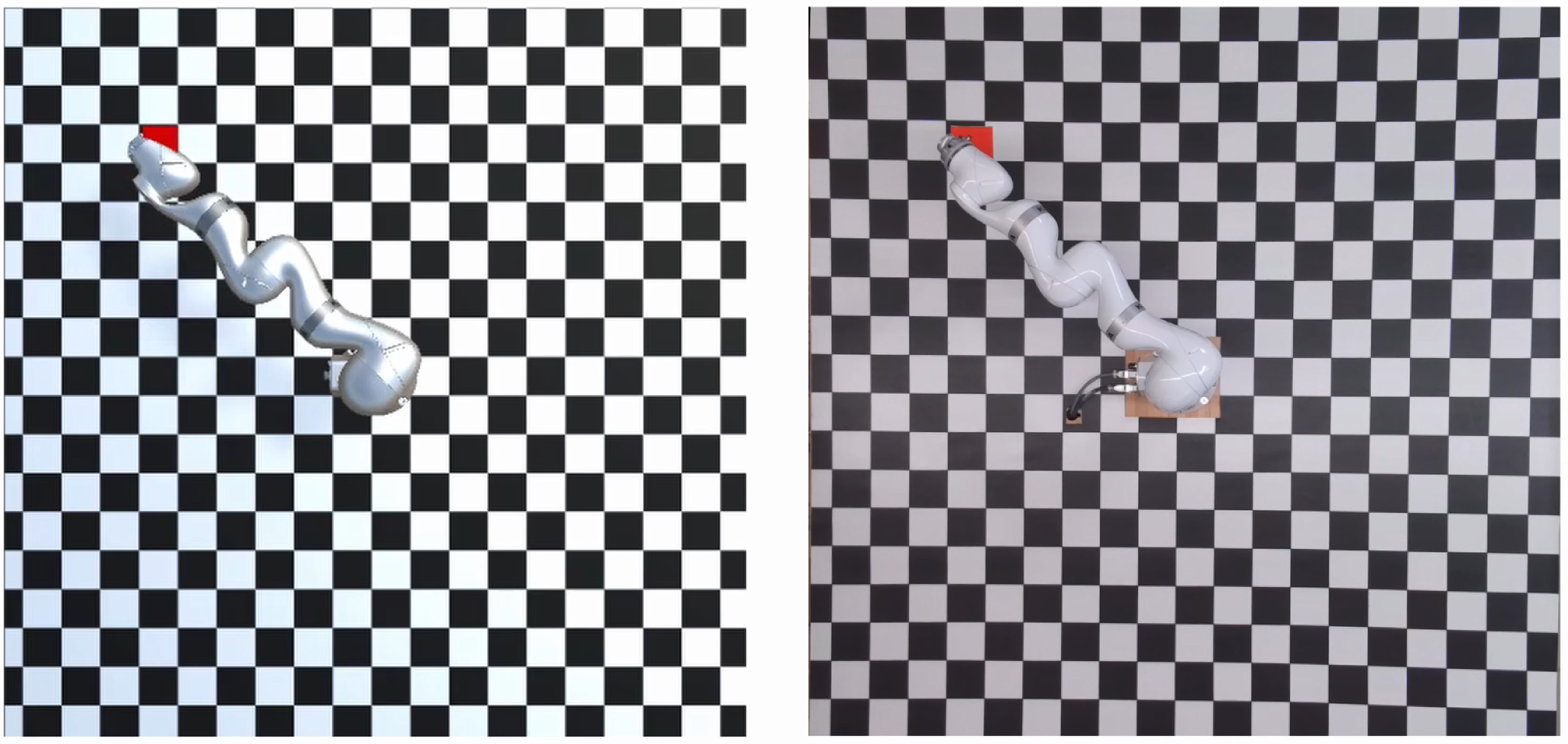}
	\caption{Left: The simulated environment. Right: The real environment.}
    \label{fig:sim_real_envs}
\end{figure}

The evaluation procedure is illustrated in Figure \ref{fig:Approach}b. While the training takes place only in simulation, the evaluation takes place both in simulation (ideal simulation setting) and on the real robot, see Figure \ref{fig:sim_real_envs}. 

The conditions for evaluating the trained model both on the simulated and the real robot are identical. Namely, the robot always starts at the `home' position, and the same 50 randomly generated target positions are used in simulation and on the real robot. 
The same maximal joint speed of $\pi/9 \; \sfrac{rad}{sec}$ ($20 \; \sfrac{deg}{sec}$) is used. 
The maximum allowed position for joint 1 is $\pm$150 degrees and $\pm$100 degrees for joint 2.
An episode duration of 500 timesteps (10 seconds) is used  to give enough time to the robot to reach and balance on the farthest targets. However, the episode can also terminate if any joint reaches the predefined safety range, 20 degrees off the corresponding mechanical limit of the joint.

\section{Experimental Results}

\begin{figure}[htbp]
	\centering
	\includegraphics[scale = 0.5]{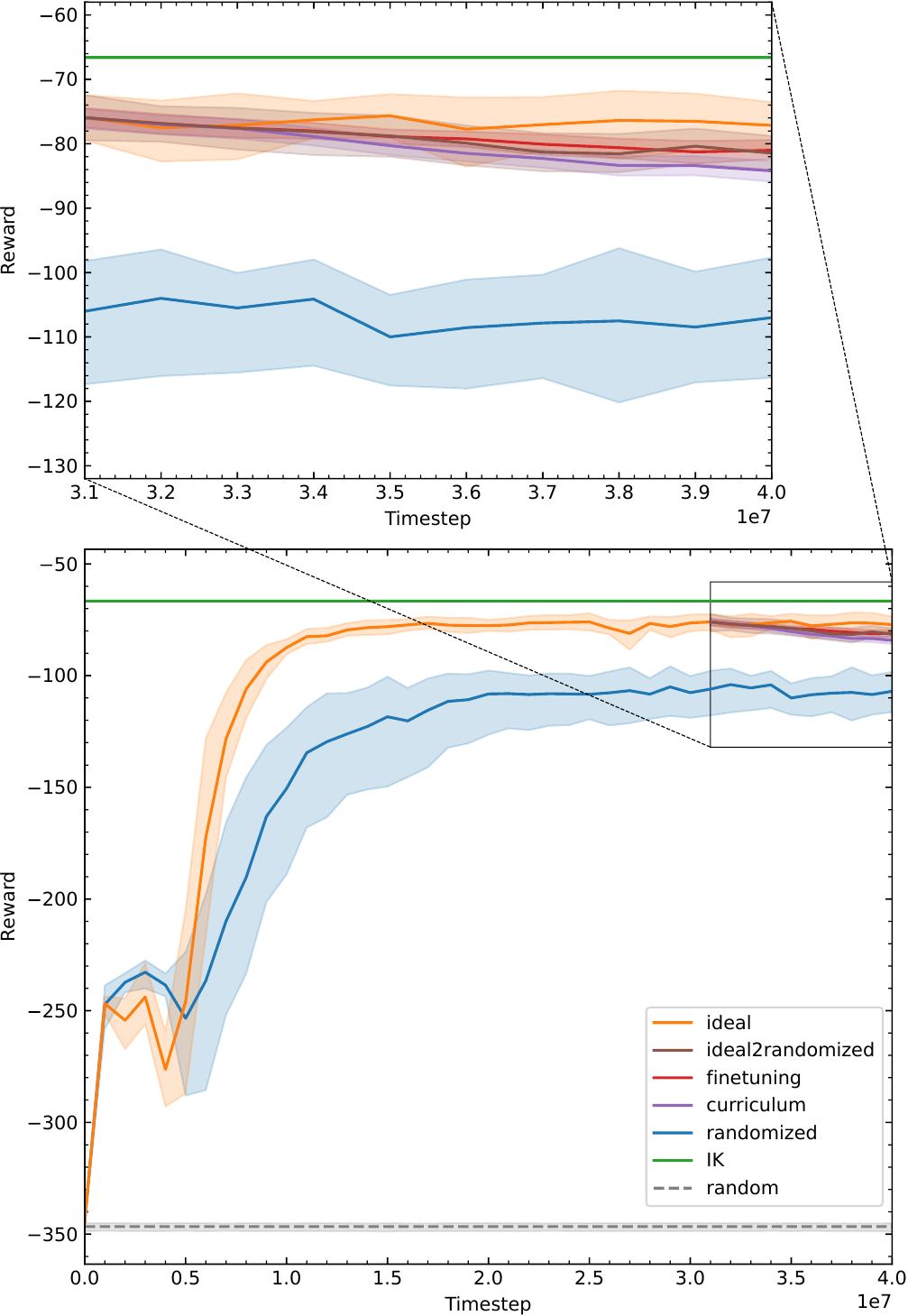}
	\caption{Training progress for $1 \; \sfrac{rad}{sec}$ measured every $10^6$ timesteps with the same random targets. The IK solver in green is the baseline. The shaded area represents 95\% confidence intervals.}
    \label{fig:training_evaluation}
\end{figure}
Figure \ref{fig:training_evaluation} shows the learning evolution of the different strategies measured as a cumulative reward. The results consist of an average of 5 different random initializations.
Except for the fully randomized strategy, all strategies have a small standard deviation and converge to a comparable value, approximately 4\% off the IK solver performance w.r.t. a random agent. 
The fully randomized strategy reaches a performance 14\% lower than the baseline and has a more substantial standard deviation between different initializations. The fully randomized strategy's improvement rate is also lower than the other strategies. 

All strategies show a performance decline early in training, starting when the agent collides with the floor. The duration of this performance decline indicates how long each strategy takes to avoid collisions. 

Table \ref{table:experimentResults} shows a summary of the results. 
Although the fully randomized strategy has the worst performance during training, it has the smallest differential between simulation and reality. Moreover, its results on the real robot are within the confidence intervals of the training results. 
Furthermore, it leads to the best performance on the real robot. 

On the contrary, the real-world performance for all other randomization strategies is substantially different from their simulation results, which the standard deviation values cannot account for. 

From the randomization strategies that start with a pre-trained model as a basis, the fine-tuning strategy consistently achieves the best or comparable performance to the other strategies on the real robot regardless of the randomization order. 
Only the TLN randomization order leads to substantially worse performance for the fine-tuning strategy, but this result is still within the range of results obtained with the other randomization strategies. 

Curriculum learning and ideal2randomized show a substantial difference in their simulation results and lead to worse real-world performance than the ideal simulation condition.

\setlength{\tabcolsep}{1.5mm}
\begin{table}[htb]
    \centering
    \caption{Mean reward for different randomization strategies.}
    \label{table:experimentResults}
    \begin{tabular}{p{16mm} cccc@{}}
    \hline\noalign{\smallskip}
       &                                     &              \specialcell{Avg.\ in\\Simulation} & \specialcell{Best Model\\(Simulation)} & \specialcell{Best Model\\(Real)} \vspace{-9pt}\\
Strategy & Seq.\ & & &\\
    \hline\noalign{\smallskip}
    ideal & N/A &  -77.14 $\pm$ 4.88  &  \textbf{-72.03} & -142.50  \\
    fine-tuning & N/A &  -81.02 $\pm$ 4.31  &  N/A  &  N/A   \\
       & $TNL$  &  -79.45 $\pm$ 1.58  &  -76.66  &  -113.82  \\
       & $TLN$  &  -83.92 $\pm$ 3.81  &  -80.38  &  -132.71  \\
       & $NTL$  &  -84.30 $\pm$ 5.91  &  -77.69  &  -114.94  \\
       & $NLT$  &  -78.58 $\pm$ 2.72  &  -74.90  &  -115.21   \\
       & $LTN$  &  -80.98 $\pm$ 2.27  &  -78.74  &  -119.74  \\
       & $LNT$  &  -78.89 $\pm$ 5.49  &  -74.55  &  -113.01  \\
    curriculum & N/A &  -84.21 $\pm$ 4.35 &  N/A  &  N/A \\
       & $TNL$  &  -84.05 $\pm$ 3.33  &  -79.85  &  -118.85  \\
       & $TLN$  &  -87.61 $\pm$ 6.09  &  -81.83  &  -139.47  \\
       & $NTL$  &  -85.55 $\pm$ 3.75  &  -79.40  &  -154.58  \\
       & $NLT$  &  -83.74 $\pm$ 5.68  &  -76.99  &  -129.13  \\
       & $LTN$  &  -82.59 $\pm$ 2.99  &  -77.67  &  -171.78  \\
       & $LNT$  &  -81.71 $\pm$ 2.42  &  -78.99  & -159.85  \\ 
    ideal2random & N/A &  -81.46 $\pm$ 3.48 &  -76.54  & -167.53 \\
    randomized & N/A & -107.00 $\pm$ 12.19  &  -96.01  &  \textbf{-109.10}  \\
\hline
    IK solver & N/A &  -66.60  &  \textbf{-66.60}  &  \textbf{-77.52}  \\    
    \hline\noalign{\smallskip}
    \end{tabular}
\end{table}

The two best randomization strategies, i.e., full randomization and fine-tuning, were further investigated.
The experimental setup was kept identical, except for the maximal joint velocities, which were decreased to $\pi/9 \sfrac{rad}{sec}$ during training to match the joint velocity during testing. The episode length was increased to 500 timesteps to allow enough time for solving the task at a lower speed.

\begin{figure}[htb]
	\centering
	\includegraphics[scale = 0.5]{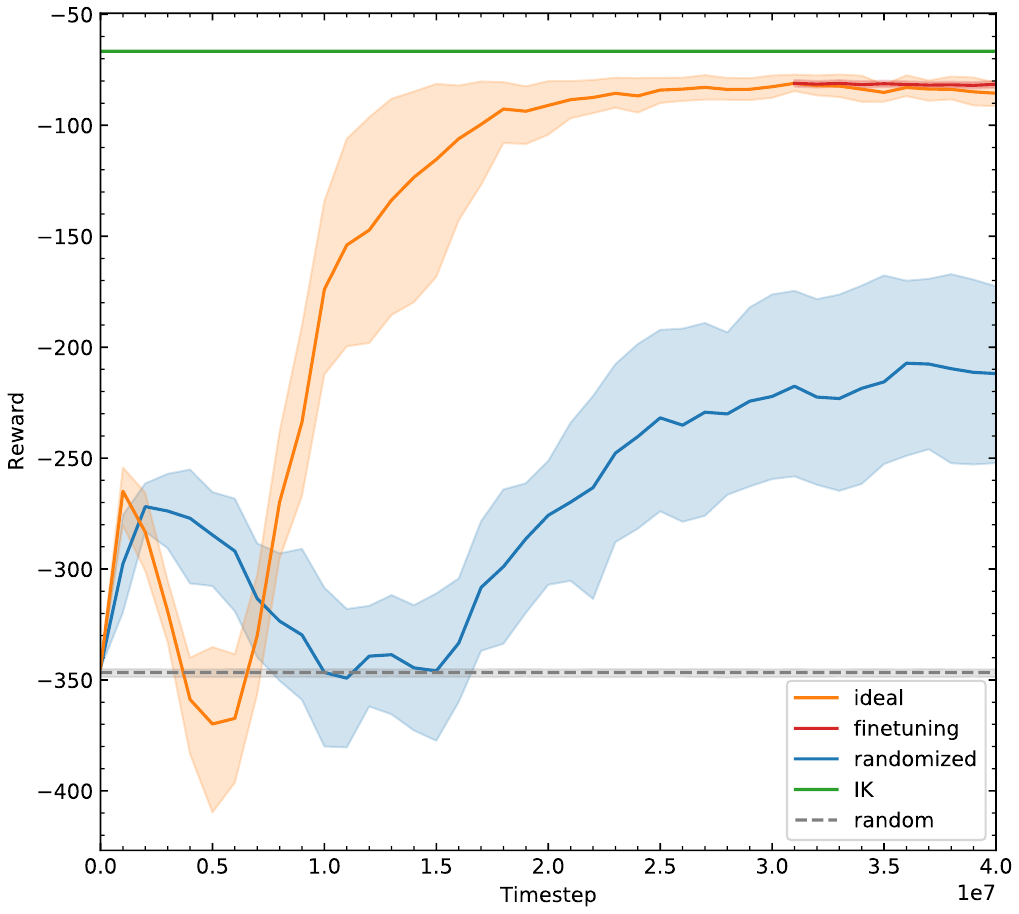}
	\caption{Training progress for $\pi/9 \sfrac{rad}{sec}$ joint speed measured every $10^6$ timesteps with the same random targets. The IK solver in green is the baseline. The shaded area represents 95\% confidence intervals.}
    \label{fig:rewards_training_20_deg}
\end{figure}

Figure \ref{fig:rewards_training_20_deg} shows that a lower joint velocity makes the problem more challenging because exploring the whole search space takes longer. 
However, the same trends as for the 1 rad/sec training condition were observed.
The fine-tuning strategy converges to a comparable value to the ideal simulation case and is approximately 5\% off the baseline w.r.t. the random agent performance. 
The fully randomized strategy converges to a substantially lower cumulative reward value of approximately 51\% off the baseline and at a slower rate than when trained at a speed of 1 rad/sec.
All the strategies also show a decline in performance early in training due to the collision to the floor. This time learning to avoid collisions takes longer than the initial setup due to the decreased joint speeds. 

Table \ref{table:experiment2Results} summarizes the results for $\pi/9 \sfrac{rad}{sec}$ joint speed in training for both in simulation and on the real robot.

\setlength{\tabcolsep}{1.5mm}
\begin{table}[htb]
    \centering
    \caption{Mean reward for the best randomization strategies under lower maximal joint speeds.}
    \label{table:experiment2Results}
    \begin{tabular}{p{13mm} cccc@{}}
    \hline
       &                                     &              \specialcell{Avg.\ in\\Simulation} & \specialcell{Best Model\\(Simulation)} & \specialcell{Best Model\\(Real)} \vspace{-9pt}\\
Strategy & Seq.\ & & &\\
    \hline
    ideal & N/A &   -84.16 $\pm$ 5.01 &  -79.05 & -129.10  \\
    fine-tuning & N/A &  -81.71 $\pm$ 6.41  &  N/A  &  N/A   \\
       & $TNL$  &  -79.29 $\pm$ 3.96  &  -74.20  &  \textbf{-115.17}  \\
       & $TLN$  &  -84.19 $\pm$ 12.14 &  -75.60  &  -125.63  \\
       & $NTL$  &  -79.05 $\pm$ 5.36  &  \textbf{-73.74}  &  -117.33  \\
       & $NLT$  &  -80.42 $\pm$ 4.08  &  -75.83  &  -121.82   \\
       & $LTN$  &  -81.76 $\pm$ 3.91  &  -78.25  &  -117.41  \\
       & $LNT$  &  -85.53 $\pm$ 5.50  &  -77.79  &   -135.18  \\
    randomized & N/A & -210.26 $\pm$ 47.85  &  -158.69  &  -197.22  \\
    \hline  
    \end{tabular}
\end{table}

The performance for the fine-tuning strategy remains practically unchanged both in simulation and on the real robot. In this case, the performance on the real robot also falls outside the confidence intervals determined in the simulations. 
However, the fully-randomized agents cannot achieve the same performance in either the simulation or the real robot. Moreover, the simulation and the real robot performance substantially deteriorate when training at a lower joint speed. 
However, the performance on the real robot still falls within the confidence intervals determined in the simulations. 
\section{Discussion}

In qualitative terms, we observed that the fully-randomized model reached and stabilized near the target location for most of the targets during the evaluation on the real robot. In contrast, the models of the other strategies showed a visible jitter around the target location, which was subjectively worse with the curriculum and the ideal2randomized models. In addition, there was a significant difference in the number of times the episode terminated due to joint limits being reached in the first experiment: 52\% for ideal2randomized, 46\% for ideal simulation, and 28.7\% for curriculum learning, while only 8\% for the fully randomized simulation and 4.33\% for the fine-tuning models. For the second experiment, on the other hand, when the training scenario was more similar to the evaluation scenario due to the same joint velocity, the model trained under ideal simulation conditions terminated the least due to reaching joint limits (10\%), followed by 16\% for randomized and 20.3\% for the fine-tuning strategy.
The qualitative differences are visible in the supplementary video material.

While the randomized strategy could find a suitable policy under the first experiment, reducing the joint speeds in combination with the unchanged randomization ranges made the exploration under fully-randomized settings significantly more challenging in the second experiment. The lower joint velocity leads to notably worse model performance both in simulation and on the real robot. This worse performance indicates that strong randomization can reduce the model convergence rate or the ability to find a suitable policy under some conditions. 

Interestingly, the fine-tuning strategy performed similarly under both joint velocity conditions. Moreover, it led to overall lower episode terminations due to safety limits when training and testing velocity conditions differed. Therefore, starting from a naive solution in ideal simulation followed by sequential randomization can be an alternative to fully-randomized training for Sim2Real transfer. Especially when the ranges are not based on system identification and might be unrealistic, or many randomization parameters are combined simultaneously.

\section{Conclusion}
We investigated the randomization effects on Sim2Real Transfer in reinforcement learning for a 2~DoFs robot reaching task. 
Four different randomization strategies were tested: fine-tuning, curriculum learning, ideal2randomized, and fully randomized simulations. 
Three commonly used parameters in Sim2Real transfer in robotics were randomized: latency, torque, and sensor and motor noise. 

Curriculum learning and ideal2randomized despite obtaining comparable results in simulation as fine-tuning and the naive ideal simulation did not transfer well onto the real robot and consistently lead to worse performance than the ideal simulation. 

Our results show that fully-randomized simulations can lead to the best real robot performance, although not always. Despite that, we believe it is a promising strategy to bridge the reality gap because the real robot performance consistently falls within the confidence interval obtained in simulations.  
However, inappropriately intensive randomization might negatively affect the overall performance, even below the ideal simulation condition. 

On the contrary, the fine-tuning strategy leads to a consistent real robot's performance regardless of randomization order or the intensity of the randomization. However, the real robot's performance cannot be anticipated based on the simulated results of the fine-tuning strategy.  

Finally, additional experiments on the real robot are necessary to determine whether these results are statistically significant. 

\paragraph{Future Work}
We plan to determine the effect size and explore the generalization of our results by including more complex robotic tasks involving contact, like pushing, grasping, or pick \& place. 
We would also like to explore the effect of design decisions on Sim2Real performance, akin to the joint velocity used for training, which can severely impact the performance of some strategies. 
We want to explore sequential randomization strategies further because they seem to lead to a more consistent real-world performance but have a more significant reality gap than fully-randomized strategies. 
Finally, we plan to work with end-to-end robotic models relying on visual input, and analyze visual domain randomization under a similar methodology.

\bibliographystyle{IEEEtran}
\bibliography{paper}

\end{document}